\documentclass[conference]{IEEEtran}

\usepackage{times}
\usepackage{epsfig}
\usepackage{graphicx}
\usepackage{subfig}

\ifCLASSINFOpdf
\else
\fi
\usepackage[linesnumbered,ruled]{algorithm2e}

\usepackage[cmex10]{amsmath}
\begin{document}
%
\title{Graph Scaling Cut with L1-Norm for Classification of Hyperspectral Images}

\author{\IEEEauthorblockN{Ramanarayan Mohanty}
\IEEEauthorblockA{Advanced Technology\\Development Centre\\
Indian Institute of Technology\\
Kharagpur, India 721302\\
Email: ramanarayan@iitkgp.ac.in}
\and
\IEEEauthorblockN{S L Happy}
\IEEEauthorblockA{Dept. of Electrical Engineering\\ 
Indian Institute of Technology\\
Kharagpur, India 721302\\
Email: happy@iitkgp.ac.in }
\and
\IEEEauthorblockN{Aurobinda Routray}
\IEEEauthorblockA{Dept. of Electrical Engineering\\ 
Indian Institute of Technology\\
Kharagpur, India 721302\\
Email: aroutray@iitkgp.ac.in }}


%


\maketitle

\begin{abstract}
In this paper, we propose an L1 normalized graph based dimensionality reduction method for Hyperspectral images, called as `L1-Scaling Cut' (L1-SC). The underlying idea of this method is to generate the optimal projection matrix by retaining the original distribution of the data. Though L2-norm is generally preferred for computation, it is sensitive to noise and outliers. However, L1-norm is robust to them. Therefore, we obtain the optimal projection matrix by maximizing the ratio of between-class dispersion to within-class dispersion using L1-norm. Furthermore, an iterative algorithm is described to solve the optimization problem. 
The experimental results of the HSI classification confirm the effectiveness of the proposed L1-SC method on both noisy and noiseless data.       
\end{abstract}



%
\IEEEpeerreviewmaketitle

\section{Introduction}
Hyperspectral remote sensing images with high spatial and spectral resolution is used to capture the inherent properties of the surface. These hyperspectral images (HSI) contains a huge number of contiguous spectral bands. It spreads over a narrow spectral bandwidth with wealth of information content. These informations are used for characterization, identification and classification of the physical and chemical properties of the land-cover with improved accuracy. The huge spectral bands implies the high dimensional redundant HSI data. This high dimensionality is the major challenge in HSI classification. In order to overcome this challenge, dimensionality reduction (DR) is usually applied to the HSI data. An effective DR method reduces the high dimension data into low dimensional representative features. This DR method improves the classification performance by reducing computational complexity and exploring the intrinsic property of the reduced data features.

In the field of HSI processing, a large number of DR approaches have been developed during past few years. Among them in unsupervised category principal component analysis (PCA) \cite{martinez2001pca} is widely used one. In addition, several supervised DR approaches have also been developed, linear discriminant analysis (LDA) \cite{ye2004optimization} is the most popular classical approach. Among LDA and PCA, LDA uses the labeled information for DR and performs better than PCA in classification task. However, LDA always assumes the data distribution as Gaussian with equal variance and unimodal. Hence, it fails to handle the real HSI data which is heteroscedastic and multimodal in nature. A graph based scaling cut (SC) \cite{zhang2009local}, \cite{zhang2015scaling} method addresses these problem by constructing the pairwise similarity matrix among the samples of the classes. 

The graph based SC method basically makes a projection of the data into a lower dimensional space by maximizing the variance of the input data points. Although, the SC method by Zhang \textit{et al.} in \cite{zhang2015scaling} works well for the multimodal hyperspectral data, they are very sensitive in handling the outliers and noise in the dataset. The conventional SC works by computing the dissimilarity matrix among the data samples. This dissimilarity matrix computation is mostly done by calculating the conventional L2-norm between the samples. The square operation in L2-norm criterion magnifies the outliers \cite{ding2006r}, \cite{wang2014fisher}. Therefore, the presence of outliers drift the projection vectors from the desired projection direction. Hence, dimension reduction and classification of hyperspectral data demands robust algorithms that are resistant to possible outliers. 

It is found that, the L1-norm based DR method is a robust alternative to handle outliers problem in image classification \cite{wang2014fisher}, \cite{liu2017non}, \cite{ke2005robust}, \cite{kwak2008principal}, \cite{li2010l1}. Kwak \textit{et.al.} \cite{kwak2008principal} computed the covariance matrix using L1-norm and proposed PCA-L1 by greedy strategy. Ke and Kanade, in \cite{ke2005robust}, proposed L1-PCA by using the alternative convex method to solve the projection matrix.  Similarly in \cite{wang2014fisher}, Wang \textit{et.al.} proposed LDA-L1 by solving the supervised LDA method using L1-norm maximization in an iterative manner. Li \textit{et.al.} \cite{li2015robust} proposed the 2D version of the LDA (L1-2DLDA) using the L1-norm optimization.  

L1-norm based LDA has achieved excellent performance for image classification \cite{wang2014fisher}\cite{li2015robust}. However, our HSI data is heteroscedastic and multimodal. SC has proved its worth \cite{zhang2015scaling} HSI classification. Motivated by these literature, we propose a L1-norm based scaling cut method (L1-SC) for DR and classification of HSI data. 
In this work we formulate the SC algorithm into an L1-norm optimization problem by maximizing the ratio of between-class dissimilarity and within-class dissimilarity matrix. Then, we solve this L1-norm optimization problem by using an iterative algorithm to generate a projection matrix. The projected reduced dimension HSI data are further used for classification by using support vector machine (SVM) classifier. We analyze the classification performance by applying it over the spectral information of two real world HSI datasets.

The rest of the paper is organized as follows. In section~\ref{sec:prelimnary}, a brief introduction to the conventional L2-norm based SC method is discussed. We present the proposed L1-SC method including its objective function and algorithmic procedure for its solution in section~\ref{sec:proposed_work}. Then section~\ref{sec:res_analysis} enumerates the experimental results of the proposed L1-SC method over two HSI datasets. Finally, we give the conclusive remarks to our work in section~\ref{sec:conclusn}.   

\section{Conventional L2- Norm based Graph Scaling Cut Criterion Revisited} \label{sec:prelimnary}
The purpose of SC is to determine the mapping matrix for projecting the original data into a lower dimension space. The classical LDA method is computed based on the assumption that the data distribution of each class is Gaussian with equal variance. However, the distribution of real world data is more complex than Gaussian. Hence LDA fails when data is heterscedastic and multimodal. The major advantage of the SC over the state-of-art LDA is handling these heteroscedastic and multimodal data. This method eleminates the Gaussian distribution limitation of LDA by constructing the dissimilarity matrix among the data samples. 

Let $X =(x_1, x_2, ...,x_n) \in R^{D\times n}$ is the input training dataset, given by $\{x_i,L_i\}|_{i=1}^n$. Here $L_i = \{1,2,...,C\}$ is the class label of the corresponding training data with total $C$ classes and $n$ training data samples. The objective is to determine a projection matrix, that project the input training data of $D$ dimensions into reduced $d$ dimension such that $d << D$. The between-class dissimilarity matrix and the within-class dissimilarity matrix of SC are defined as
\begin{equation}
\begin{split}
S_{B_k}^{SC}\, = \,\sum\limits_{x_i\, \in \,{U_k}} {\sum\limits_{x_j\, \in \,{{\bar U}_k}} {\frac{1}{{{n_k}{n_{\bar{k}}}}}\,({x_i} - {x_j}){{({x_i} - {x_j})}^T}} }\\
S_{W_k}^{SC}\, = \,\sum\limits_{x_i\, \in \,{U_k}} {\sum\limits_{x_j\, \in \,{{U}_k}} {{\frac{1}{{{n_k}{n_k}}}}({x_i} - {x_j}){{({x_i} - {x_j})}^T}} }  
\end{split}
\end{equation}
where $U_k$ represents all the samples from $k$th class and $n_k$ is the total number of elements in $U_k$. Similarly, $\bar{U_k}$ represents all the data points that that does not belong to the $k$th class and ${n_{\bar{k}}}$ denotes the total number of elements in $\bar{U_k}$. $S_{B_k}^{SC}$ represents the dissimilarity between $U_k$ class and $\bar{U_k}$, whereas $S_{W_k}^{SC}$ is the dissimilarity matrix within the $U_k$ class. Based on the $S_{W_k}^{SC}$ and $S_{B_k}^{SC}$, the objective function of SC can be written as
\begin{align}\label{eq: scaling_cut}
\begin{aligned}
Scut(W)\, & = \,\frac{{\left| {\sum\limits_{k\, = \,1}^c {{W^T}{S_{B_k}^{SC}}W} } \right|}}{{\left| {\sum\limits_{k\, = \,1}^c {({W^T}{S_{W_k}^{SC}}W\, + {W^T}{S_{B_k}^{SC}}W\,)} } \right|}}\\
& = \,\frac{{\left| {{W^T}{S_{B}^{SC}}W} \right|}}{{\left| {{W^T}{({S_{W}^{SC}} + {S_{B}^{SC}})W}} \right|}}\\
&= \,\frac{{\left| {{W^T}{S_{B}^{SC}}W} \right|}}{{\left| {{W^T}{S_{T}^{SC}}\,W} \right|}}
\end{aligned}
\end{align}
where $S_{B}^{SC}=\sum\limits_{k=1}^C{S_{B_k}^{SC}}$, $S_{W}^{SC}=\sum\limits_{k=1}^C{S_{W_k}^{SC}}$, and $S_{T}^{SC} = (S_{B}^{SC} + S_{W}^{SC})$ is the total dissimilarity matrix and $W$ is the projection matrix. The dissimilarity matrix is scaled according to the size of the class. Hence, this graph cut is termed as scaling cut.

\section{Proposed L1-Norm based Scaling Cut Criterion} \label{sec:proposed_work}
The conventional L2-norm based graph scaling cut criterion basically determines the projection matrix by maximizing the between-class distances, and minimizing the within-class distances to enhance the compactness among the data points. These models characterize the geometric structure of the data by computing the L2-norm. These L2-norm is computed by using square euclidean distance, which is sensitive to outliers and noises \cite{wang2014fisher}, \cite{liu2017non}. These outlier elements drift the projection vectors from the desired projection directions. Hence, it reduces the flexibility of L2-norm based algorithms. To handle this issue, L1-norm based technique is widely used as a robust alternative of conventional L2-norm based technique. Motivated by the idea of L1-norm based modeling, we propose to model the graph based scaling cut criterion by using the L1-norm optimization instead of L2-norm optimization. This L1-norm based SC method is solved by following iterative algorithm

\subsection{L1-norm based Graph Scaling Cut ($L1-SC$)}
Inspired by the existing literatures on L1-norm based method \cite{liu2017non}, \cite{li2015robust}, we propose to maximize the SC criterion using L1-norm rather than L2-norm. The equation (\ref{eq: scaling_cut}) can be simplified to a trace ratio \cite{fukunaga2013introduction} problem, which can further be reduced to the Frobenius norm, given by,
\begin{align}
&W^* = \max_{W^TW = I} \frac{Tr(W^TS_{B}^{SC}W)}{Tr(W^TS_{T}^{SC}W)} \nonumber \\
&= \max_{W^TW = I}\frac{\sum\limits_{\substack{k; x_i \in U_k;\\x_j \in \bar{U}_k}}\frac{1}{n_k n_{\bar{k}}} Tr\Big(W^T({x_i} - {x_j}){{({x_i} - {x_j})}^T}W\Big)}{\sum\limits_{\substack{k; x_i \in U_k;\\x_j \in U_k}} \frac{1}{n_k n_{k}}  Tr\Big(W^T({x_i} - {x_j}){{({x_i} - {x_j})}^T}W\Big)} \nonumber \\
&= \max_{W^TW = I}\frac{\sum\limits_{k = 1}^c\sum\limits_{x_i \in U_k}\sum\limits_{x_j \in \bar{U}_k} \frac{1}{n_k n_{\bar{k}}} \left\|W^T({x_i} - {x_j})\right\|_F^2}{\sum\limits_{k = 1}^c\sum\limits_{x_i \in U_k}\sum\limits_{x_j \in U_k} \frac{1}{n_k n_{k}} \left\|W^T({x_i} - {x_j})\right\|_F^2}
\end{align}
As can be observed, the above objective is based on Frobenius norm, which also involves the square operations and in term, it is sensitive to outliers to noise and outliers similar to L2-norm. In order to reduce the sensitivity, we use the objective function in terms of the L1-norm. The proposed model of the objective function for L1-norm SC is defined as,


\begin{align} \label{eq:l1sc_optmiz_prblm}
v_{opt} &= \max_{v^Tv = 1}
\frac{\sum\limits_{k\, = \,1}^c\sum\limits_{x_i \in U_k}\sum\limits_{x_j \in \bar{U}_k} \left\|{{{v^T}\,{\frac{1}{{{n_k}{n_{\bar{k}}}}}\,({x_i} - {x_j})}} } \right\|_1}{\sum\limits_{k\, = \,1}^c \sum\limits_{x_i \in U_k}\sum\limits_{x_j \in {U}_k} \left\| {{{{v^T}\,{\frac{1}{{{n_k}{n_{k}}}}\,({x_i} - {x_j})}}} } \right\|_1} \nonumber \\
&= \max_{v^Tv = 1}
\frac{\sum\limits_{k = 1}^c\sum\limits_{x_i \in U_k}\sum\limits_{x_j \in \bar{U}_k} \frac{1}{{{n_k}{n_{\bar{k}}}}} |{{{v^T}{({x_i} - {x_j})}} }|}{\sum\limits_{k = 1}^c \sum\limits_{x_i \in U_k}\sum\limits_{x_j \in {U}_k} \frac{1}{{{n_k}{n_{k}}}} |{{{{v^T}{({x_i} - {x_j})}}} }|}
\end{align} 

The objective of the criterion (\ref{eq:l1sc_optmiz_prblm}) is to find the optimal projection vector $v$ that maximize the ratio of between-class dispersion to the within-class dispersion. These optimized projection vectors are used to construct the optimal projection matrix $ V = \{v_1, v_2, ..., v_d\}$. These projecting vectors are sequentially optimized in $d$ directions. 
We derive the following iterative algorithm to find the optimal projection vector $v$ that maximizes the objective function (\ref{eq:l1sc_optmiz_prblm}). The entire algorithmic procedure for L1-SC method is listed below.

\subsection{Algorithmic Procedure for L1-SC}
The aforementioned objective function (\ref{eq:l1sc_optmiz_prblm}) involves maximization of L1-norm based optimization problem. We solve this problem by an iterative algorithm to obtain the optimal projection vector $v^*$ of the matrix $V$.

The objective function (\ref{eq:l1sc_optmiz_prblm}) seems similar to the trace ratio formulation of the general graph scaling cut in \cite{zhang2009local} and \cite{zhang2015scaling}.
It is difficult to solve (\ref{eq:l1sc_optmiz_prblm}) by the traditional optimization techniques as both numerator and denominator are constructed by L1-norm maximization and minimization. Inspired by the idea used in \cite{wang2014fisher}, \cite{kwak2008principal}, \cite{li2015robust} and \cite{wang2012l1}, we are using similar L1-norm optimization technique in this work. Thus, we solve the objective function (\ref{eq:l1sc_optmiz_prblm}) to find the optimal projection vector $v^*$ by the iterative technique. The algorithmic procedure of L1-SC is given as follows.  

\begin{itemize}
\item[1.]  The iteration variable $t$ is set to zero ($t=0$). Then we randomly initialize the $d$ dimensional vector $v(t)$ and normalize it such that ${v(t)}^T{v(t)} = 1$.
\item[2.] Two sign functions are defined to compensate the absolute value operation for the numerator and denominator term of (\ref{eq:l1sc_optmiz_prblm}) . These sign functions are computed as
\begin{equation}\label{eq:sign_func}
\begin{split}
{q_{ij}}(t) = & \,\left\{ {\begin{array}{*{20}{c}}
{1,\,\,\,\,\,\,\mbox{if}\,\,\,\,{v^T}(t)({x_i} - {x_j})\,\, > 0}\\
{ - 1,\,\,\,\,\,\mbox{if}\,\,\,{v^T}(t)({x_i} - {x_j})\, \le 0}
\end{array}} \right.\\
 & \mbox{and} \\
 {r_{ij}}(t) = & \,\left\{ {\begin{array}{*{20}{c}}
{1,\,\,\,\,\,\,\mbox{if}\,\,\,\,{v^T}(t)({x_i} - {x_j})\,\, > 0}\\
{ - 1,\,\,\,\,\,\mbox{if}\,\,\,{v^T}(t)({x_i} - {x_j})\, \le 0}
\end{array}} \right.
\end{split}
\end{equation}
\item[3.] Use the sign function to compute $p(t)$ and $b(t)$ by the following equation:
\begin{equation}\label{eq:pt_bt}
\begin{split}
p(t) = & \sum\limits_{k = 1}^c \sum\limits_{i = 1}^{{n_k}}{\sum\limits_{j = 1}^{{n_{\bar{k}}}} {{q_{ij}}(t){\frac{1}{{{n_k}{n_{\bar{k}}}}}\,{v^T}(t)({x_i} - {x_j})}} } \\
b(t) = & \sum\limits_{k = 1}^c \sum\limits_{i = 1}^{{n_k}}{\sum\limits_{j = 1}^{{n_k}} {{r_{ij}}(t){\frac{1}{{{n_k}{n_{k}}}}\,{v^T}(t)({x_i} - {x_j})}} } \\ 
\end{split}
\end{equation}
Then using $p(t)$ and $b(t)$, update $g(v(t))$
\begin{equation} \label{eq:find_g_vt}
g(v(t)) = \frac{p(t)}{{v(t)}^T{p(t)}}-\frac{b(t)}{{v(t)}^T{b(t)}} 
\end{equation}
\item[4.] Then update the vector $v(t)$ using $g(v(t))$ by
\begin{equation}\label{eq:update_vt}
v(t+1) = v(t) + \gamma g(v(t))
\end{equation}
where $\gamma$ is the learning rate parameter (a small positive value). Then normalize the $v(t+1)$ and update $t=t+1$. If any denominator in (\ref{eq:find_g_vt}) happen to zero then perturb the $v(t)$ with a small non zero random vector $\Delta v$ and update it by $v(t) = (v(t)+\Delta v)/{||(v(t)+\Delta v)||}$ and start with step-2. 
\item[5.] Convergence check: If the $v(t)$ doesn't show significant increment or $||v(t+1)-v(t)|| \le \epsilon$ or total iteration number is greater then maximum given iteration number, then go to step-2 otherwise go to step-6.
\item[6.] Stop iteration and assign $v^* = v(t)$. 
\end{itemize}

Above procedure only gives one optimal projection vector. In practical classification problem this one vector is not sufficient for the projection. Hence, It need a projection matrix consists of multiple projection vectors placed in its column space to optimize the objective function. These projection vectors are used to update the input data matrix by
\begin{equation}
X \leftarrow X - {v^*}({v^*}^T)X
\end{equation}
and then the projection matrix $V$ is padded as $V=[V,{v^*}]$.

Using the above procedure, we can form the optimal projection matrix $V$ of size $R^{D \times d}$. The pseudo-code for the complete algorithmic procedure for the projection matrix of L1-SC is listed in Algorithm~$1$.
\begin{algorithm} \label{alg:L1_SC}
	\SetKwInOut{Input}{Input}
	\SetKwInOut{Output}{Output}
	\SetKwInOut{Result}{Result}
	\Input{The training dataset $ \{x_i, L_i\} _{i = 1}^n \in R^{D \times n}$; \\ $L_i$ is the label of each training data $x_i$; \\ Desired dimensionality is $d$ and $d \ll D$.}
	Formulate the L1-norm based objective function in (\ref{eq:l1sc_optmiz_prblm}) to solve the optimization problem.\\
	Determine the optimal projection vector $v^*$ by solving the optimization problem (\ref{eq:l1sc_optmiz_prblm}) in Algorithm~2 \\
    Update the input data by using $X = X - {v^*}{v^*}^TX$. \\
    Pad these optimal projection vectors $v^*$ into the optimal matrix by $V=[V,{v^*}]$. \\ 
	Project the original data into the lower dimensional space $d$ by projection matrix $V$  \\
	\Output {Projection matrix $V=\{v_1, v_2, ..., v_d\}$ $\in R^{D \times d}$, consists of $d$ projection vectors}
	\Result{Projected matrix $Y = {V^T}X$}
	\caption{L1-norm based scaling cut algorithm}
\end{algorithm}

\begin{algorithm} \label{alg:projc_mat_L1-SC}
	\SetKwInOut{Input}{Input}
	\SetKwInOut{Output}{Output}
	\SetKwInOut{Result}{Result}
    \SetKwProg{For}{For}{}{}

    \Input{
    Number of projection vector $d$ ($d \ll D$); \\ Learning rate parameter $\gamma$; \\ Maximum number of iteration is \textit{itmax}}
    Set $t=0$ and Initialize $v(0)$ to a $D$ dimensional random vector such that ${v(0)}^T{v(0)}=1$ \\
        Compute the sign function $q_{ij}(t)$ and $r_{ij}(t)$ using (\ref{eq:sign_func}) and set $p(t)$ and $b(t)$ 		using (\ref{eq:pt_bt})\\
        Determine the $g(v(t))$ function using (\ref{eq:find_g_vt}) to update the $v(t)$.\\ 
        Update the $v(t)$ by using (\ref{eq:update_vt}). where $\gamma > 0$ is the learning parameter.\\
        Converge if: $||v(t+1)-v(t)|| \le \epsilon$ or $t > itmax$
	 
	\Output {Projection vector $v^* = v{(d)}$}
	\caption{Computation of projection vector for $v{^*}$}
\end{algorithm}

\section{Experimental Results and Analysis}\label{sec:res_analysis}
In this section we evaluate the performance of the proposed L1-SC method on two HSI datasets\footnote{http://www.ehu.eus/ccwintco/index.php?title=Hyperspectral\_Remote\\\_Sensing\_Scenes}: Salinas $(D=204, C=16)$ and Pavia center $(D=102, C=9)$. Then we compare it with the state-of-art conventional LDA \cite{ye2004optimization}, SC \cite{zhang2015scaling}, LSC \cite{zhang2013semisupervised} and L1-LDA \cite{wang2014fisher}. In conventional L2-norm based methods use PCA as preprocessing but in L1-norm methods we don't use any preprocessing step. In classification stage, we use SVM classifier with linear kernel to identify the robustness of the proposed algorithm.
\begin{figure}[h]
	\centering
    \subfloat[]{
	\label{subfig:Salinas_comp}
	\includegraphics[width=0.48\textwidth]{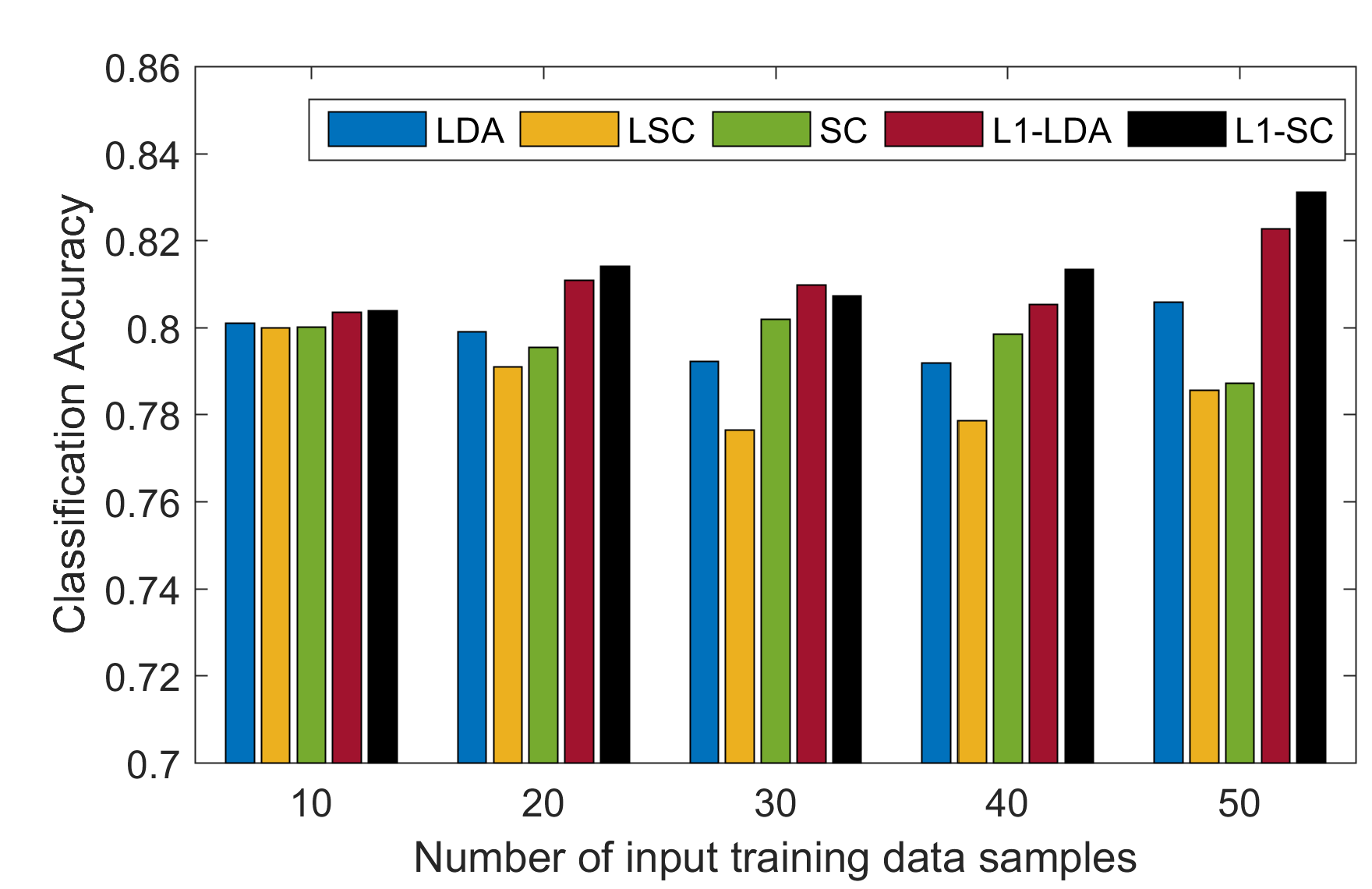} } 
 
\subfloat[]{
	\label{subfig:Pavias_comp}
	\includegraphics[width=0.48\textwidth]{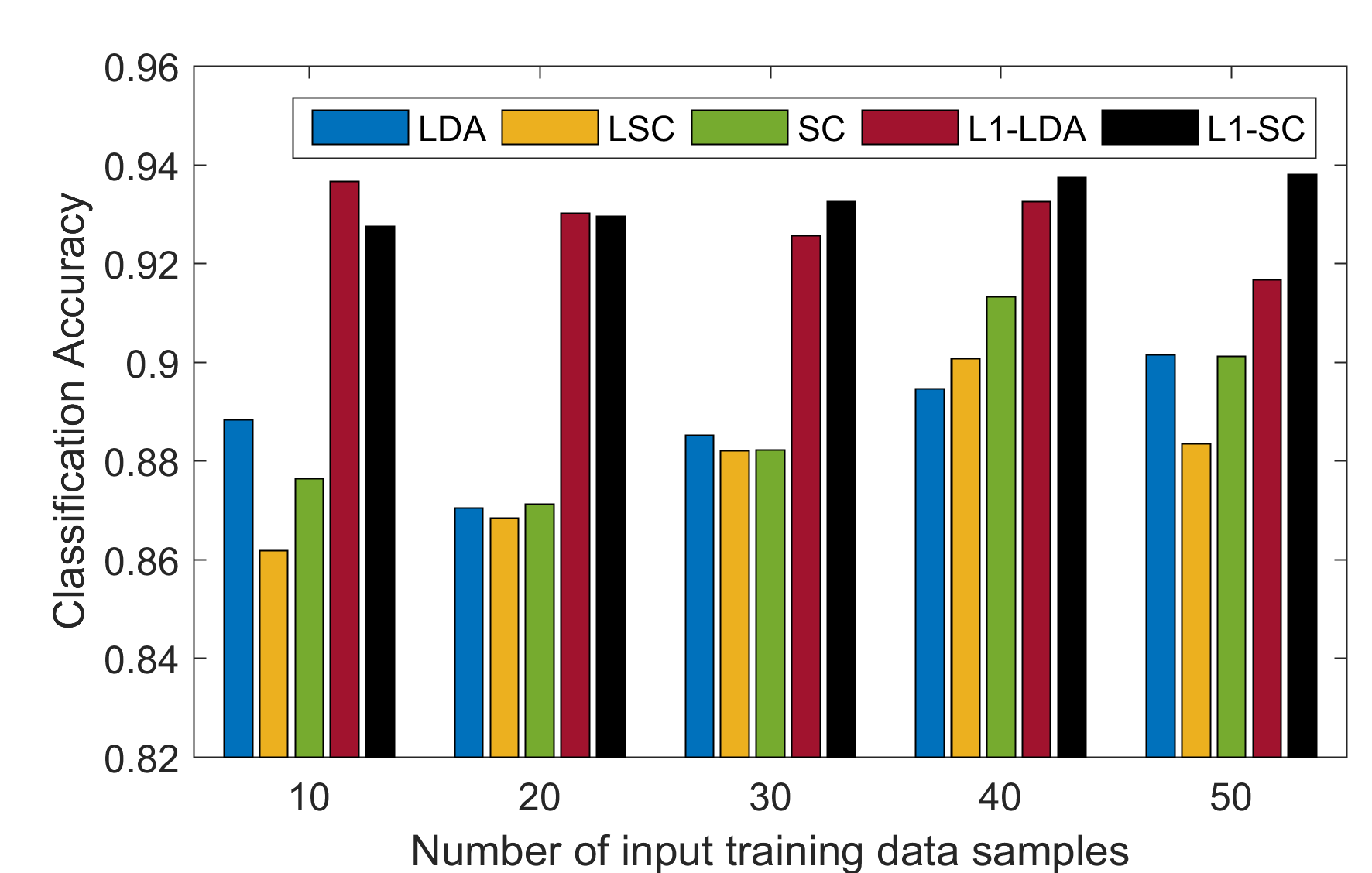} } 
    
\caption{ Effects of different number input samples for the methods on overall accuracies on two data sets Salinas (\ref{subfig:Salinas_comp}) and Pavia center (\ref{subfig:Pavias_comp}). From left to right of X-axis shows the overall accuracies with $10, 20, 30, 40$ and $50$ number input data samples for $10$ dimensions.}
	\label{fig:salina_pavia_comp}
\end{figure}

\begin{table*}[h]
\centering
\caption{Classification performance of proposed approach compared with other L2-norm and L1-norm based approaches }
\label{tab:Classfcn_L1_l2}
\begin{tabular}{l|c c c|c c c}
\hline
\multicolumn{1}{c|}{Dataset} & \multicolumn{3}{c|}{\textbf{Salinas}}                                 & \multicolumn{3}{c}{\textbf{Pavia Center}}                                   \\ \hline
Methods                       & Accuracy + stdv       					& F1-Score              	& Dims                  & Accuracy + Stdv       & F1-Score              & Dims                  \\  \hline
LDA   \cite{ye2004optimization}                        &           $83.14 \pm 1.93$            &           $0.8917$            &          $35$             &          $92.34 \pm 0.64$   &   $0.8406$      &      $25$       \\ 
SC    \cite{zhang2015scaling}                       &            $81.38 \pm 1.41$            &            $0.8558$           &        $40$               &          $93.61 \pm 0.68$   &  $0.8600$       &     $25$            \\ 
LSC    \cite{zhang2013semisupervised}                       &             $83.09 \pm 1.88$          &          $0.8939$             &      $50$                 &          $93.43 \pm 1.07$   &   $0.8587$      &      $45$         \\  \hline
L1-LDA   \cite{wang2014fisher}                     &      $83.21 \pm 1.85$                &          $0.8913$             &        $30$             &        $93.50 \pm 0.61$               &         $0.8542$      &   $40$  \\ 
\textbf{L1-SC}    \textbf{(Proposed)}          &  { $\mathbf{84.01 \pm 1.67}$ } & {$\mathbf{0.8956}$} & {$\mathbf{15}$} & {$\mathbf{94.20 \pm 0.63}$} & {$\mathbf{0.8724}$} & {$\mathbf{10}$} \\ \hline
\end{tabular}
\end{table*}


In these experiments, we randomly select $10$ training samples from each class of the dataset and rest of the samples are used as test dataset. All obtained results are the average of the $5$ iterations. Here we evaluate and analyze the effectiveness of the proposed L1-SC by determining the overall classification accuracy of the SVM classifier on the projected data.

Fig.~\ref{fig:salina_pavia_comp} shows the behavior of different L2-norm and L1-norm based algorithms in terms of overall classification accuracy with respect to varied number of input samples. Here, Fig.~\ref{subfig:Salinas_comp} and Fig.~\ref{subfig:Pavias_comp} shows the overall classification accuracy of Salinas and Pavia dataset for input data size from $10$ to $50$. From this figure, it is clearly observed that the proposed L1-SC method completely outperforms the L2-norm based methods and performs on par with L1-LDA when the input data sample size is less.
\begin{figure}[h]
	\centering
    \subfloat[]{
	\label{subfig:Salinas_Ns_comp}
	\includegraphics[width=0.48\textwidth]{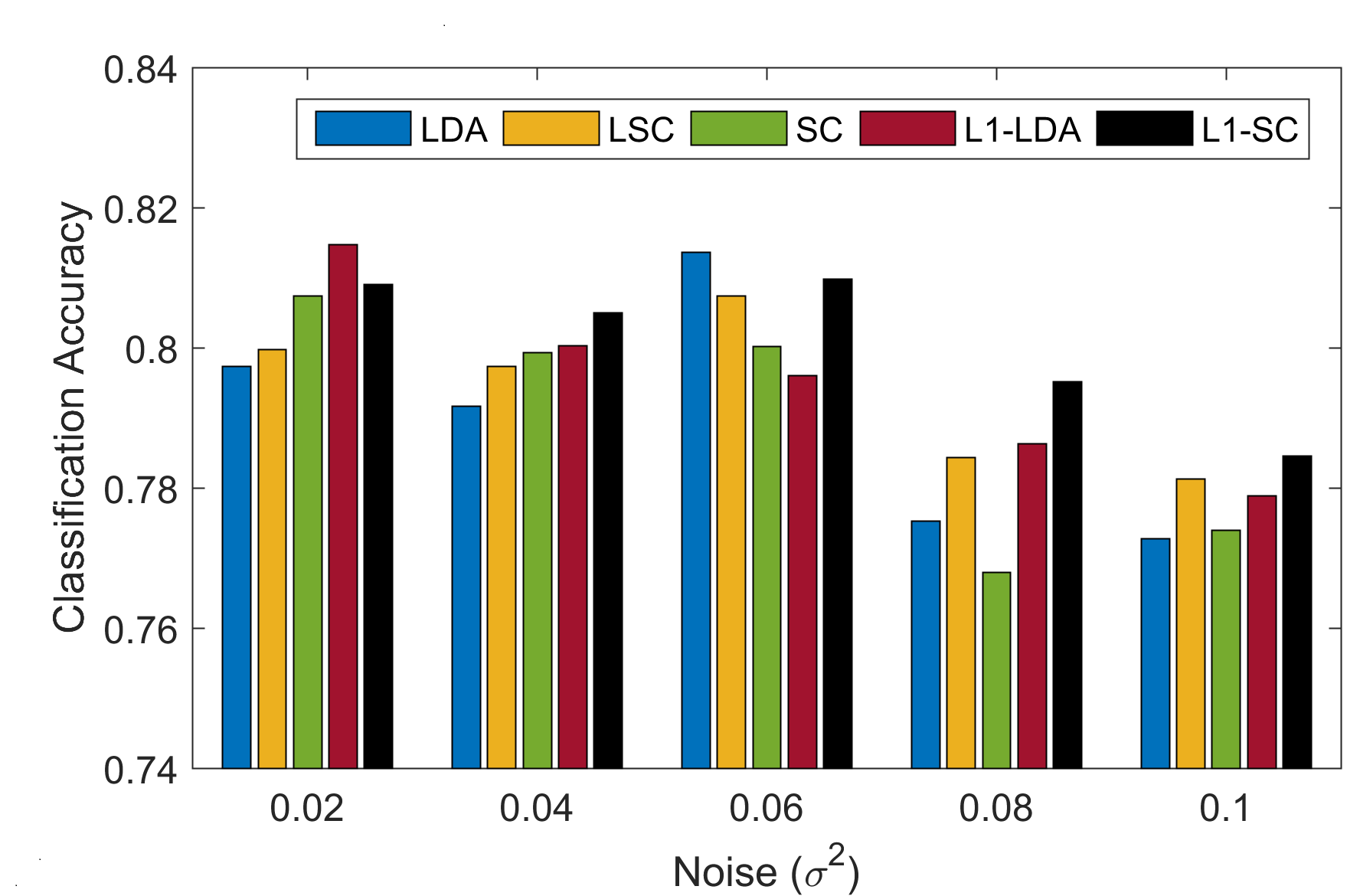} } 
 
\subfloat[]{
	\label{subfig:Pavias_NS_comp}
	\includegraphics[width=0.48\textwidth]{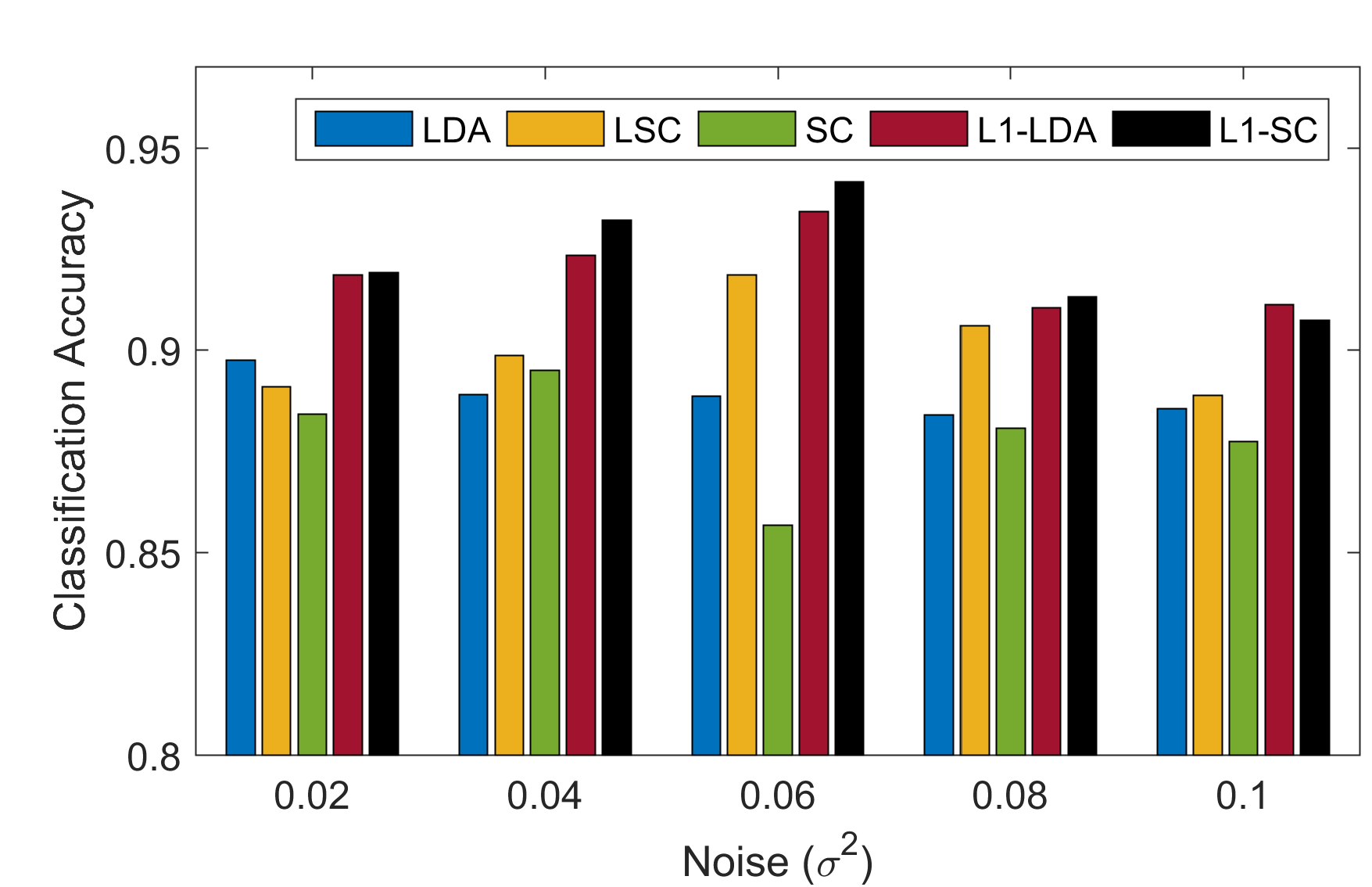} } 
    
\caption{ Illustration of noise robustness of the proposed method with respect to other methods on two data sets Salinas (\ref{subfig:Salinas_Ns_comp}) and Pavia center (\ref{subfig:Pavias_NS_comp}).}
	\label{fig:salina_pavia_noise_comp}
\end{figure}

To better illustrate the noise robustness feature of the proposed L1-SC method with respect to others, we inject white Gaussian noises of different levels to the raw input HSI data and performed the classification on these data using different approaches. The variance of the noise level is varied from 2\% to 10\% of the variance of the pixel values. As Fig.~\ref{fig:salina_pavia_noise_comp} indicate, the proposed L1-SC method is more robust to noises and achieve better classification accuracies than other methods.   

The proposed L1-SC method is compared with other popular L2-norm and L1-norm based methods. The statistics of the highest overall classification accuracy along with corresponding $F1$-score and dimension of the algorithms for Salinas and Pavia center dataset are highlighted in Table~\ref{tab:Classfcn_L1_l2}. Here in Table~\ref{tab:Classfcn_L1_l2}, all the results taken classification accuracy are the average of $5$ runs using $10$ random training samples per class. In order to show the robustness of the algorithm, we have considered the $F1$-score along with the overall classification accuracy as the performance measure \cite{sokolova2009systematic}. Table~\ref{tab:Classfcn_L1_l2} gives the following observations
\begin{itemize}
\item The overall classification accuracy of L1-norm based methods performs better than the other state-of-art L2-norm based methods.
\item The classification results of the proposed L1-SC method outperforms the other L2-norm based methods and L1-LDA for both the datasets with less dimensions.
\item In Salinas dataset, the proposed L1-SC method produces highest accuracy with maximum $F1$-score among other approaches by considering only $15$ dimensions. Similarly in case of Pavia center dataset, it takes only $10$ dimensions. This shows the effectiveness of the algorithm in finding proper projection direction. 
\end{itemize}
The above observations clearly explains the robustness of the proposed algorithm in low dimensional feature space.

\section{Conclusion}\label{sec:conclusn}
In this study, we have proposed a novel DR method L1-SC by computing the L1-norm based inter-class and intra-class dispersion. This method determines the projection directions by exploiting the discriminant structure and preserving the geometrical structure of the data. Our method differs from other state-of-art in various ways. For instance, it preserves the intrinsic property as well as the distribution of the data and we believe it handles multimodal and heteroscedastic data with noise and outliers quite well. We examined the performance of our method and other methods over two real world HSI datasets. The promising results of L1-SC on these two datasets demonstrates its noise robustness and efficiency. 







\bibliographystyle{IEEEtran}
\bibliography{mybib}

\begin{thebibliography}{10}
\providecommand{\url}[1]{#1}
\csname url@samestyle\endcsname
\providecommand{\newblock}{\relax}
\providecommand{\bibinfo}[2]{#2}
\providecommand{\BIBentrySTDinterwordspacing}{\spaceskip=0pt\relax}
\providecommand{\BIBentryALTinterwordstretchfactor}{4}
\providecommand{\BIBentryALTinterwordspacing}{\spaceskip=\fontdimen2\font plus
\BIBentryALTinterwordstretchfactor\fontdimen3\font minus
  \fontdimen4\font\relax}
\providecommand{\BIBforeignlanguage}[2]{{%
\expandafter\ifx\csname l@#1\endcsname\relax
\typeout{** WARNING: IEEEtran.bst: No hyphenation pattern has been}%
\typeout{** loaded for the language `#1'. Using the pattern for}%
\typeout{** the default language instead.}%
\else
\language=\csname l@#1\endcsname
\fi
#2}}
\providecommand{\BIBdecl}{\relax}
\BIBdecl

\bibitem{martinez2001pca}
A.~M. Mart{\'\i}nez and A.~C. Kak, ``Pca versus lda,'' \emph{Pattern Analysis
  and Machine Intelligence, IEEE Transactions on}, vol.~23, no.~2, pp.
  228--233, 2001.

\bibitem{ye2004optimization}
J.~Ye, R.~Janardan, C.~H. Park, and H.~Park, ``An optimization criterion for
  generalized discriminant analysis on undersampled problems,'' \emph{Pattern
  Analysis and Machine Intelligence, IEEE Transactions on}, vol.~26, no.~8, pp.
  982--994, 2004.

\bibitem{zhang2009local}
X.~Zhang, S.~Zhou, and L.~Jiao, ``Local graph cut criterion for supervised
  dimensionality reduction,'' in \emph{Sixth International Symposium on
  Multispectral Image Processing and Pattern Recognition}.\hskip 1em plus 0.5em
  minus 0.4em\relax International Society for Optics and Photonics, 2009, pp.
  74\,962I--74\,962I.

\bibitem{zhang2015scaling}
X.~Zhang, Y.~He, L.~Jiao, R.~Liu, J.~Feng, and S.~Zhou, ``Scaling cut
  criterion-based discriminant analysis for supervised dimension reduction,''
  \emph{Knowledge and information systems}, vol.~43, no.~3, pp. 633--655, 2015.

\bibitem{ding2006r}
C.~Ding, D.~Zhou, X.~He, and H.~Zha, ``R 1-pca: rotational invariant l 1-norm
  principal component analysis for robust subspace factorization,'' in
  \emph{Proceedings of the 23rd international conference on Machine
  learning}.\hskip 1em plus 0.5em minus 0.4em\relax ACM, 2006, pp. 281--288.

\bibitem{wang2014fisher}
H.~Wang, X.~Lu, Z.~Hu, and W.~Zheng, ``Fisher discriminant analysis with
  l1-norm,'' \emph{IEEE transactions on cybernetics}, vol.~44, no.~6, pp.
  828--842, 2014.

\bibitem{liu2017non}
Y.~Liu, Q.~Gao, S.~Miao, X.~Gao, F.~Nie, and Y.~Li, ``A non-greedy algorithm
  for l1-norm lda,'' \emph{IEEE Transactions on Image Processing}, vol.~26,
  no.~2, pp. 684--695, 2017.

\bibitem{ke2005robust}
Q.~Ke and T.~Kanade, ``Robust l/sub 1/norm factorization in the presence of
  outliers and missing data by alternative convex programming,'' in
  \emph{Computer Vision and Pattern Recognition, 2005. CVPR 2005. IEEE Computer
  Society Conference on}, vol.~1.\hskip 1em plus 0.5em minus 0.4em\relax IEEE,
  2005, pp. 739--746.

\bibitem{kwak2008principal}
N.~Kwak, ``Principal component analysis based on l1-norm maximization,''
  \emph{IEEE transactions on pattern analysis and machine intelligence},
  vol.~30, no.~9, pp. 1672--1680, 2008.

\bibitem{li2010l1}
X.~Li, Y.~Pang, and Y.~Yuan, ``L1-norm-based 2dpca,'' \emph{IEEE Transactions
  on Systems, Man, and Cybernetics, Part B (Cybernetics)}, vol.~40, no.~4, pp.
  1170--1175, 2010.

\bibitem{li2015robust}
C.-N. Li, Y.-H. Shao, and N.-Y. Deng, ``Robust l1-norm two-dimensional linear
  discriminant analysis,'' \emph{Neural Networks}, vol.~65, pp. 92--104, 2015.

\bibitem{fukunaga2013introduction}
K.~Fukunaga, \emph{Introduction to statistical pattern recognition}.\hskip 1em
  plus 0.5em minus 0.4em\relax Academic press, 2013.

\bibitem{wang2012l1}
H.~Wang, Q.~Tang, and W.~Zheng, ``L1-norm-based common spatial patterns,''
  \emph{IEEE Transactions on Biomedical Engineering}, vol.~59, no.~3, pp.
  653--662, 2012.

\bibitem{zhang2013semisupervised}
X.~Zhang, Y.~He, N.~Zhou, and Y.~Zheng, ``Semisupervised dimensionality
  reduction of hyperspectral images via local scaling cut criterion,''
  \emph{IEEE Geoscience and Remote Sensing Letters}, vol.~10, no.~6, pp.
  1547--1551, 2013.

\bibitem{sokolova2009systematic}
M.~Sokolova and G.~Lapalme, ``A systematic analysis of performance measures for
  classification tasks,'' \emph{Information Processing \& Management}, vol.~45,
  no.~4, pp. 427--437, 2009.

\end{thebibliography}
%



\end{document}